\documentclass{article}

\usepackage[final, nonatbib]{neurips_2019}
     
\usepackage{times}
\usepackage{soul}
\usepackage{url}
\usepackage[hidelinks]{hyperref}
\usepackage[utf8]{inputenc}
\usepackage[small]{caption}
\usepackage{graphicx}
\usepackage{amsmath}
\usepackage{booktabs}
\usepackage{algorithm}
\usepackage{color}
\usepackage{amsfonts}
\usepackage{mathtools}
\usepackage{amsthm}
\usepackage{algpseudocode}
\usepackage{subcaption}
\usepackage{tabularx}
\usepackage[numbers]{natbib}

\DeclareMathOperator*{\argmax}{argmax} 
\DeclarePairedDelimiter{\abs}{\lvert}{\rvert}

\DeclareMathOperator*{\Expectation}{E} 
\newcommand{\mathbbm}[1]{\text{\usefont{U}{bbm}{m}{n}#1}} 
\algdef{SE}[DOWHILE]{Do}{doWhile}{\algorithmicdo}[1]{\algorithmicwhile\ #1}%

\newtheorem{proposition}{Proposition}
\title{Unifying Ensemble Methods for Q-learning\\via Social Choice Theory}

\author{
    Rishav Chourasia\thanks{Rishav Chourasia performed this research during an internship at MPI-SWS.}\\
    National University of Singapore\\
    Singapore \\
    \texttt{rishav1@comp.nus.edu.sg} \\
    \And
    Adish Singla \\
    Max Planck Institute for Software Systems (MPI-SWS) \\
    Saarbrücken, Germany \\
    \texttt{adishs@mpi-sws.org} \\
}

\begin{document}

\maketitle


\begin{abstract}
\looseness-1
Ensemble methods have been widely applied in Reinforcement Learning (RL) in order to enhance stability, increase convergence speed, and improve exploration. These methods typically work by employing an aggregation mechanism over actions of different RL algorithms. We show that a variety of these methods can be unified by drawing parallels from committee voting rules in Social Choice Theory. We map the problem of designing an action aggregation mechanism in an ensemble method to a voting problem which, under different voting rules, yield popular ensemble-based RL algorithms like Majority Voting Q-learning or Bootstrapped Q-learning. Our unification framework, in turn, allows us to design new ensemble-RL algorithms with better performance. For instance, we map two diversity-centered committee voting rules, namely Single Non-Transferable Voting Rule and Chamberlin-Courant Rule, into new RL algorithms that demonstrate excellent exploratory behavior in our experiments.

\end{abstract}
\section{Introduction}


Ensemble methods such as bagging or boosting are used to improve several Machine Learning predictors \cite{zhang2012ensemble} in a manner that decreases bias and variance or improves accuracy. In Reinforcement Learning (RL), ensemble methods have been used in a variety of ways for achieving stability \cite{2016arXiv161101929A}, robustness to perturbation \cite{2016arXiv161001283R} or for improving exploration \cite{NIPS2016_6501}. One popular method is to maintain a set of quality (Q) function estimates that serves as an approximation to the Q-functions' posterior distribution, conditioned on interaction history \cite{Osband2013}. Quite similar to bagging \cite{opitz1999popular}, these Q-function estimates are trained on different state sequences in order to represent a close sample from the actual Q-function distribution \cite{NIPS2016_6501} which, upon aggregation, reduces variance. The mechanism used to aggregate the estimates can affect the algorithm's characteristics considerably \cite{Wiering:2008}.

Alternate aggregation mechanisms are employed by different ensemble RL algorithms: Majority Voting and Rank Voting Q-learning \cite{Fausser2015,DBLP:journals/corr/HarutyunyanBVN14} use ordinal aggregation mechanisms that depend on the order, rather than value, of Q-function estimates; on the other hand, Boltzmann Addition, Boltzmann Multiplication, and Averaged Q-learning \cite{Marivate:2013:ELC:2908286.2908312,Wiering:2008} use cardinal aggregation mechanisms. While a few empirical comparisons among these varied techniques have been done \cite{Wiering:2008}, any broad unification of these variants in order to identify key differences that cause significant deviation in learning characteristics hasn't been attempted to the best of our knowledge. Based on recent work on representing committee voting rules via \textit{committee scoring functions} \cite{Elkind2017JustifiedRI}, we attempt this unification in the paper and, in process, port properties of election rules to RL algorithms.





By viewing the task of aggregating Q-function estimates for action decision as a multi-winner voting problem from Social Choice Theory (SCT), we present an abstraction to several ensemble RL algorithms based on an underlying voting rule. We consider individual units (we call heads) in ensemble as the voters, action choices as the candidates, and the heads' Q-values over the actions as the ballot preference of respective voters. With this perspective on the setting, the paper has the following contributions.

\begin{itemize}
    \item We develop a generic mechanism based on multi-winner voting \cite{Elkind2017JustifiedRI} in SCT that exactly emulates the aggregation method for several of the aforementioned algorithms, when plugged with an appropriate committee scoring function. 
    \item We map four popular multi-winner voting rules---Bloc rule, Borda rule, Single Non-Transferable Voting (SNTV) \cite{norris2004electoral}, and Chamberlin-Courant rule (CCR) \cite{chamberlin1983representative}---to new ensemble RL algorithms, among which SNTV and CCR demonstrate an improvement in exploration.
\end{itemize}


We believe that our mapping method helps correlate properties of voting rules to the characteristics of respective RL algorithms. For instance, our experiments suggest a strong correlation between the \textit{Proportional Representation} (PR) \cite{bogdanor1984proportional} property of a voting rule and the consequent RL algorithm's \textit{exploration} ability.

\section{Background}
\label{sec:background}

Here, we provide a brief introduction to RL problem, describe some ensemble-based RL algorithms, and discuss committee voting strategies. For brevity, we denote $\{1,2,...,n\}$ by $[n]$ and uniform distribution over a set $A$ as $Unif(A)$. Some notations are overloaded between Section \ref{sec:ensemble} and \ref{sec:committee-voting} to aid understanding.

\subsection{Preliminaries}
\label{sec:preliminaries}

We consider a typical RL environment as a Markov Decision Process (MDP) $M := (S, A, T, R, \gamma)$, where $S$ is the state space, $A$ is the action space, and $\gamma \in [0,1)$ is the discount factor. $R(s,a)$ gives the finite reward of taking action $a$ in state $s$. $T(s'|s,a)$ gives the environment's transition probability to state $s'$ on taking action $a$ in $s$. We denote the space of all bounded real functions on $S \times A$ as $\mathbb{Q} := \mathbb{Q}_{S,A}$. A policy $\pi$ can be a deterministic mapping over actions, i.e. $\pi: S \rightarrow A$, or a stochastic distribution over the same. Every policy $\pi$ has a corresponding $Q$ function $Q^\pi \in \mathbb{Q}$ computed from the following recursive equation:
\begin{equation}
	\label{eq:q-function}
	Q^\pi(s,a) = R(s,a) + \gamma \Expectation_{\substack{s' \sim T(\cdot| s, a) \\ a' \sim \pi(s')}} [Q^\pi(s', a')].
\end{equation}



An optimal policy $\pi^*$ is defined as a policy mapping that receives maximum expected discounted returns on $M$, i.e. 

\begin{equation}
	\label{eq:v-function}
	\pi^* = \argmax_{\pi'} \Expectation_{\pi', T} \big[ \sum_{t=0}^{\infty} \gamma^t R(s_t, a_t) \big].
\end{equation}



Vanilla Q-learning algorithm starts with a random Q-function estimate $Q_0$ and follows an off-policy exploratory strategy to generate observation tuples $<\!\!\!s_t,a_t,r_t,s_{t+1}\!\!\!>$. The tuples are used to improve the estimate following the update
\begin{equation}
    \begin{aligned}
    \label{eq:q-learning-update}
    Q_{t+1}(s_t,a_t) &= (1 - \alpha) Q_{t}(s_t, a_t) \\
                     &+ \alpha(r_t + \gamma \max_{a' \in A} Q_{t}(s_{t+1},a')),
    \end{aligned}
\end{equation}
where $\alpha \in (0,1]$ is the learning rate. 
A popular exploratory policy is the $\epsilon$-soft policy over step $t$ greedy policy $\pi_t(s) = \argmax_{a \in A} Q_t(s,a)$.

\subsection{Ensemble Q-learning Algorithms}
\label{sec:ensemble}

Ensemble Q-learning algorithms employ a set of Q-function estimates for making action decisions. For an ensemble with $k$ heads, we denote the Q-function estimates at time $t$ as $Q^{i}_t \in \mathbb{Q} \quad \forall i \in [k]$. While the update rule for each estimate remains same as \eqref{eq:q-learning-update}, the action aggregation strategy is what differs for different ensemble algorithms.

\textbf{Majority Voting Q-learning} strategy selects the action with highest votes, where every ensemble head votes for it's greedy action (ties broken arbitrarily). 
	\begin{equation}
		\label{eq:mv-qlearning}
		\pi^{MV}_t(s) = \argmax_{a \in A} \sum_{i \in [k]} \mathbbm{1}\{ a = \argmax_{a' \in A}  Q^{i}_t(s,a') \} 
	\end{equation}

In \textbf{Rank Voting Q-learning}, every head supplies it's integer preference, $pref_i|A \rightarrow [\abs{A}]$, on actions based on the Q-function estimates in an order preserving manner. The action with the highest cumulative preference is selected. 
	\begin{equation}
		\label{eq:rv-qlearning}
		\pi^{RV}_t(s) = \argmax_{a \in A} \sum_{i \in [k]} pref_i(a)
	\end{equation}

\textbf{Average Q-learning} \cite{Fausser2015} uses the average of the ensemble Q-function estimates to decide the greedy action.
	\begin{equation}
		\label{eq:a-qlearning}
		\pi^{Avg}_t(s) = \argmax_{a \in A} \frac{1}{k}\sum_{i \in [k]} Q^i_t(s,a) 
	\end{equation}

Similar to \textbf{Bootstrapped DQN} \cite{NIPS2016_6501}, the Q-learning variant samples a head from the ensemble per episode and uses it's greedy action to make decisions. Let $\psi(t) \in [k]$ give the bootstrap head sampled for the episode that time-step $t$ falls in. 
	\begin{equation}
		\label{eq:b-qlearning}
		\pi^{Boot}_t(s) = \argmax_{a \in A} Q^{\psi(t)}_t (s,a)
	\end{equation}

\textbf{Boltzmann Addition Q-learning} \cite{Wiering:2008} averages the Boltzmann probabilities of actions over the ensemble and uses it as the distribution for action choices. 
	\begin{equation}
		\label{eq:ba-qlearning}
		\pi^{BA}_t(a|s) = \frac{1}{k} \sum_{i \in [k]} \frac{e^{Q^i_t(s,a)}}{\sum_{a' \in A} e^{Q^i_t(s,a')}} 
	\end{equation}

\subsection{Committee Voting}
\label{sec:committee-voting}

\looseness-1
Committee voting in Social Choice Theory \cite{aziz2017justified,DBLP:journals/corr/ElkindFSS15} deals with scenarios where societies, represented by set $K=[k]$, need to elect a representative committee $W$ among candidates $A=\{a_1, a_2, ...,a_m\}$. Every voter $i \in K$ has a preference order over the candidates, denoted by a utility function $\mu^i | A \rightarrow \mathbb{R}$. For a voter $i$, let $pos^i|A \rightarrow [\abs{A}]$ be a function that maps candidates to their ranks based on utility function $\mu^i$, from higher utility to lower. \citet{DBLP:journals/corr/ElkindFSS15} classified several ordinal voting rules as \textit{committee scoring rules} which can be succinctly represented by a \textit{committee scoring function} $f$ that maps utility-committee pairs to a real number. Even for some cardinal voting rules, scoring function can be used to describe them. Following is a list of such rules with respective scoring functions.


\textbf{Plurality} or Single Non-Transferable Voting \cite{grofman1999elections} rule uses a scoring function that returns $1$ if the committee $W$ has voter $i$'s most preferred candidate, otherwise $0$. Let $\alpha_{l}^i(a) = \mathbbm{1}\{ pos^i(a) \leq l\}$.
\begin{equation}
	\label{eq:plurality}
	f_{plurality}(\mu^i, W) = \max_{a \in W} \alpha^i_1(a)
\end{equation}

\textbf{Bloc} system's \cite{ball1951bloc} scoring function returns the number of candidates in the committee $W$ that are among voter $v$'s top $\abs{W}$-preferred candidates.
\begin{equation}
	\label{eq:bloc}
	f_{bloc}(\mu^i, W) = \sum_{a \in W} \alpha^i_{\abs{W}}(a)
\end{equation}

\textbf{Chamberlin-Courant} rule \cite{chamberlin1983representative} depends on a satisfaction function $\beta^i(a) = \abs{A} - pos^i(a)$, which is an order-preserving map over the utility values $\mu^i$. The rule's scoring function outputs the maximum satisfaction score from a candidate in the committee.
\begin{equation}
	\label{eq:chamberlin-courant}
	f_{ccr}(\mu^i, W) = \max_{a \in W} \beta^i(a)
\end{equation}

\textbf{Borda} rule's \cite{de1953memoire} scoring function uses \textit{Borda score} which, for a voter $i$, assigns same value as satisfaction $\beta^i(a)$. The committee's score with respect to a voter is just the sum of individual scores.
\begin{equation}
	\label{eq:broda}
	f_{borda}(\mu^i, W) = \sum_{a \in W} \beta^i(a)
\end{equation}

\textbf{Majority Judgment} rule's \cite{felsenthal2008majority} scoring function is a cardinal voting rule that outputs the sum of utility values for the candidates in the committee. 
\begin{equation}
	\label{eq:majority-judgement}
	f_{judge}(\mu^i, W) = \sum_{a \in W} \mu^i(a)
\end{equation}

\textbf{Lottery} rule or Random Ballot \cite{amar1984choosing} is a stochastic voting rule where a voter is randomly selected and it's most preferred candidate is elected. Let $i'$ be a random voter, i.e. $i' \sim Unif(K)$. Then for a masking function $\phi$, defined as $\phi(i) = \mathbbm{1}\{i = i'\}$, the scoring function with respect to a voter is the masked greedy utility.
\begin{equation}
	\label{eq:lottery}
	f_{lottery}(\mu^i, W) = \phi(i) \max_{a \in W}  \mu^i(a)
\end{equation}

Given an election pair $E = <\!\!\!K, A\!\!\!>$, committee size $n$, and a committee scoring rule $f$, the election result $\mathcal{M}_f(E, n)$ \cite{DBLP:journals/corr/ElkindFSS15} is given by 

\begin{equation}
	\label{eq:committee-voting}
	\mathcal{M}_f(E, n) = \argmax_{\substack{W \subset A \\ \abs{W}=n}} \sum_{i \in K} f(\mu^i, W),
\end{equation}
and winning score $\mathcal{G}_f(E, n)$ is the maximum score. Voting rules may or may not follow certain desirable properties such as \textit{committee monotonousity},  \textit{proportional representation} (PR), \textit{consistency}, etc. In \eqref{eq:committee-voting}, we break ties to preserve \textit{committee monotonousity} whenever possible, i.e. we try to ensure $\mathcal{M}_f(E, l) \subset \mathcal{M}_f(E, l+1)$ for all $l \in [\abs{A} - 1]$.





While the election result for Plurality, Borda, and Bloc rules can be computed by polynomial time greedy algorithm, the result for Chamberlin-Counrant rule has been shown to be NP-complete to compute \cite{Procaccia2008}. We can however get approximate results within a factor of $1-1/e$ for fixed $n$ through greedy algorithms \cite{tyler2011}.


\section{Unification Framework}
\label{sec:agg-mech}

The intuition behind the unification is the similarity between the voting rules and action aggregation mechanism method in RL ensembles. Let's consider ensemble heads $i \in K$ as voters that, on perceiving a state $s_t$ at time $t$, cast preferences over the action set $A$ in form of their Q-function estimates or a softmax over it, i.e.

\begin{align}
\label{eq:ensemble-utility}
	\mu^i(a) &= Q^i_t(s_t,a), \ \text{or} \\
\label{eq:ensemble-utility-softmax}
	\mu^i(a) &= e^{Q^i_t(s_t,a)} \big/ \sum_{a' \in A}e^{Q^i_t(s_t,a')}.
\end{align}



 \begin{figure}[!hbp]
    \includegraphics[width=0.55\textwidth]{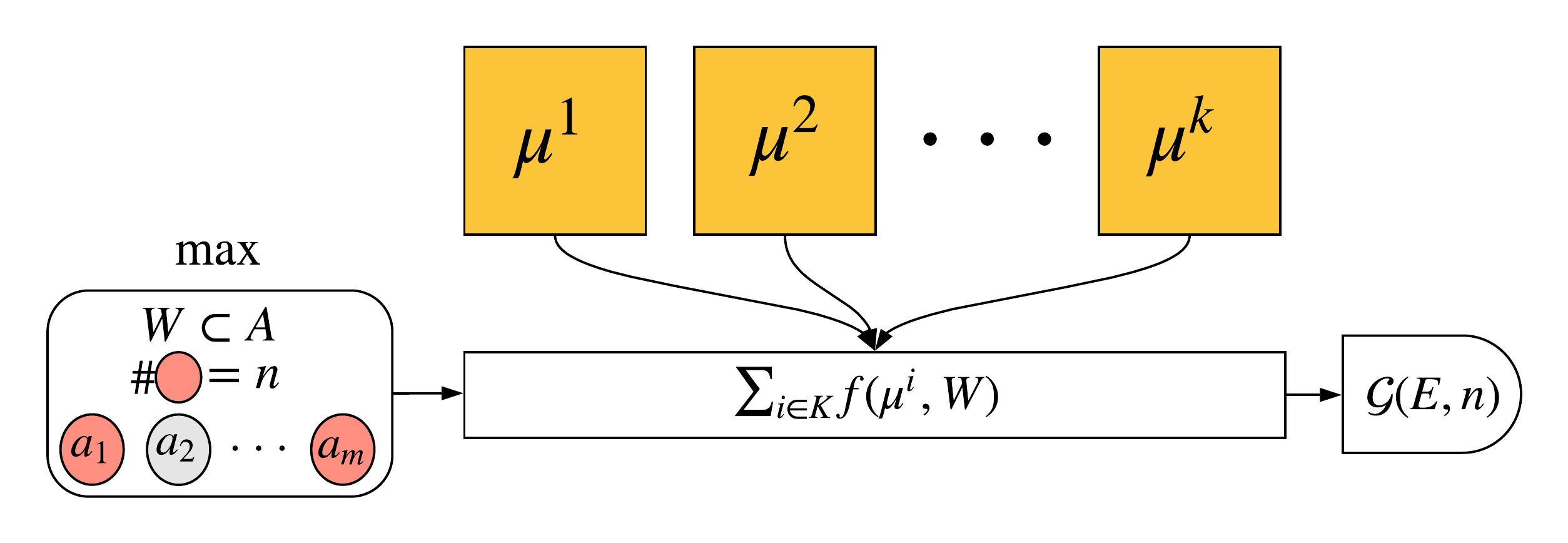}
    \centering
    \caption{\label{fig:aggregation-mech}Voters' utilities are indicated by coloured squares. For a committee $W$, scoring function $f$ returns the satisfaction score for a voter $i$ (based on it's utility $\mu^i$). Winning score $\mathcal{G}$ is the sum of individual scores, maximized over all $n$-sized committees.}
\end{figure}

For the sake of simplicity, we have omitted inputs $t, s_t$ in the utility function notation $\mu^i$ as they are identical across the ensemble for each decision reconciliation instance. Applying committee voting rules described in Section \ref{sec:committee-voting} on these utility functions is identical to several ensemble algorithms' action selection method described in Section \ref{sec:ensemble}. Figure \ref{fig:aggregation-mech} describes this aggregation mechanism.

Following propositions establish equivalence between voting rules on election pair $E$ with voter utilities given by Q-function estimates \eqref{eq:ensemble-utility} and action reconciliation in several ensemble Q-learning algorithms.

\begin{proposition}
\label{prop:equivalence}
Plurality and Bloc voting rules map to Majority Voting Q-learning; Chamberlin-Courant and Borda voting rules map to Rank Voting Q-learning; Majority Judgment voting rule maps to Average Q-learning. In brief,
\begin{equation}
	\begin{aligned}
	    &\pi^{MV}_t(s_t)  \equiv \mathcal{M}_{f_{bloc}} (E, 1) \equiv \mathcal{M}_{f_{plurality}} (E, 1), \\
	    &\pi^{RV}_t(s_t)  \equiv \mathcal{M}_{f_{ccr}} (E, 1) \equiv \mathcal{M}_{f_{borda}} (E, 1), \\
	    &\pi^{Avg}_t(s_t) \equiv \mathcal{M}_{judge} (E, 1).
	\end{aligned}
\end{equation}
\end{proposition}

Even in case of Boltzmann Addition Q-learning, that uses a stochastic policy, the probabilities of the actions can be represented using the committee voting aggregation mechanism as stated by the following proposition.

\begin{proposition}
\label{prop:equivalence-boltzmann}
Majority Judgment voting rule on softmax utility \eqref{eq:ensemble-utility-softmax}, gives Boltzmann Addition Q-learning's policy, i.e., if ties are broken in a manner that preserves \textit{committee monotonicity} and $f$ = $f_{judge}$, then for action $a_l$ defined as  

\begin{equation}
	    a_l = \mathcal{M}_{f} (E, l) \setminus \mathcal{M}_{f} (E, l-1),
\end{equation}
for an integer $l \in [n]$, the Boltzmann probabilities for $a_l$ is
\begin{equation}
	    \pi^{BA}_t(a_l | s_t) =  ( \mathcal{G}_{f} (E, l) - \mathcal{G}_{f} (E, l-1) ) \big/ \abs{K}.
\end{equation}
\end{proposition}

While action selection strategy in Bootstrapped Q-learning is quite similar to Lottery rule as the decisions in both are made based on a sampled head / ballot's utility, they differ in the sampling frequency. Lottery voting samples a new ballot every election, but Bootstrapped Q-learning uses a sampled agent for multiple steps (episode). This can be accounted for in the mechanism by modifying the masking function $\phi(i)$ used in $f_{lottery}$ \eqref{eq:lottery} to use bootstrap head $\psi(t)$ (same throughout an episode) for ongoing episode at time-step $t$.

\begin{proposition}
\label{prop:bootstrapped}
Lottery voting rule maps to Bootstrapped Q-learning when ballots are drawn randomly every episode, i.e. if $\phi(i) = \mathbbm{1}\{ i = \psi(t)\}$, where $\psi(t)$ gives the sampled head for episode at $t$, then

\begin{equation}
	\pi^{Boot}_t(s_t) \equiv \mathcal{M}_{f_{lottery}} (E, 1).
\end{equation}
\end{proposition}
\section{Multi-winner RL Algorithms}

With the exception of Boltzmann Addition Q-learning, all the RL algorithms discussed so far mapped to committee voting rules with committee-sizes restricted to $n=1$. A single candidate doesn't satisfactorily represent a community of voters. These RL algorithms, therefore, can be improved by extending them to allow multi-winner action committees. Unfortunately, the modelling of typical RL paradigm forbids selection of multiple actions at a state. Some environments do support backtracking, i.e. allow the agent to retrace to a state multiple times in order to try multiple actions \cite{de2018multi}. However, we instead follow a straightforward approach to randomly sample an action from the winning committee, i.e.


\begin{equation}
    \label{eq:sampling-action}
	\pi_t(a | s_t) = \frac{1}{n}, \quad if\ a \in \mathcal{M}_f(E,n).
\end{equation}

This uniform sampling promotes diversity in action selection as all the winning candidate actions receive equal chance of selection regardless of voter's backing.


\subsection{Dynamic Committee Resizing}
\label{sec:dyn-resizing}

Deciding a suitable committee-size for the multi-winner extension to ensemble RL algorithms is very crucial. Since every policy decision involves conducting an election over action choices, a static committee size cannot hoped to be optimal for all the observations throughout learning. We therefore propose \textbf{dynamic committee resizing} based on a satisfaction threshold hyper-parameter $S_{thresh}$, that elects committees varying in sizes across elections. Described as ELECTION procedure in Algorithm \ref{alg:ensemble-Q-learning}, this subroutine is similar to the classic heuristic analyzed by \cite{nemhauser1978analysis} for optimizing submodular set functions. We start with an empty committee and populate it iteratively with greedy candidate actions that best improves the satisfaction score until the threshold is reached or all candidates are included.

Using this subroutine, we extend the classic Q-learning to a generic ENSEMBLE Q-LEARNING algorithm described in Algorithm \ref{alg:ensemble-Q-learning}. For $S_{thresh} = 0$, this procedure mimics existing ensemble algorithms when executed with respective scoring functions mentioned in Section \ref{sec:agg-mech} and extends to novel ensemble RL algorithms with interesting properties for non-zero $S_{thresh}$.




\begin{algorithm}[!htbp]
\caption{Ensemble Q-Learning Algorithms}
\label{alg:ensemble-Q-learning} 
\begin{algorithmic}[1]
    \Procedure{Election}{$f$,$S_{thresh}$,$E$,$\mu^{(\cdot)}$}
        \State $W = \phi, S=0$
        \Do
            \State $a^* \leftarrow \argmax_{a \in A \setminus W} \sum_{i \in K} f(\mu^i, W \cup \{a\})$
            \State $W = W \cup \{a^*\}$
            \State $S = \sum_{i \in K} f(\mu^i, W)$
        \doWhile{$W \neq A$ and $S \leq S_{thresh}$}
        \State \textbf{return} $W$
    \EndProcedure

    \Procedure{Ensemble Q-learning}{$f$, $S_{thresh}, K, A$}
    \State Initialize $Q^{i}_0 \ \forall \ i \in K$ randomly.
    \State Get start state $s_0$.
    \For{$t = \{0,1,2,..\}$}
        \State Formulate utilities $\forall i \in K$: $\mu^i(a) \leftarrow Q^{i}_t(s_t, a)$.
        \State $W_t \leftarrow ELECTION(f, S_{thresh}, <\!\!\!K,A\!\!\!>, \mu^{(\cdot)})$.
        \If {t is an $\epsilon$-exploration step} 
            \State Sample $a_t \leftarrow Unif(A)$
        \Else
            \State Sample $a_t \leftarrow Unif(W_t)$
        \EndIf
        \State Play $a_t$ and receive $r_t, s_{t+1}$.
        \State Perform standard Q-estimate updates \eqref{eq:q-learning-update}.
    \EndFor
    \EndProcedure
\end{algorithmic}
\end{algorithm}

\subsection{Properties of Voting Rules in RL}

A wide variety of axiomatic studies for multi-winner voting \cite{DBLP:journals/corr/ElkindFSS15,faliszewski2017multiwinner} rules have been conducted which provide analysis of several properties, such as \textit{consistency}, \textit{solid coalitions}, \textit{unanimity}, etc. When voting rules are mapped to an ensemble RL algorithm via the aforementioned procedure, these properties affect the characteristics of the algorithm. Therefore, analyzing the effect of voting properties on their RL counterpart helps in understanding existing algorithms as well as proposing improved ones. Effective exploration in RL algorithms is a highly desirable characteristic that has been well studied \cite{thrun1992efficient} for improving sample complexity of any given task. Using our generic unified ensemble RL algorithm, we performed a study on the correlation between efficient exploration and \textit{proportional representation} voting property for various single and multi-winner voting rules. While there are several PR centric voting rules such as proportional approval voting, reweighted approval voting \cite{aziz2017justified}, and Moore rule \cite{monroe1995fully}, we limited ourselves to \textit{committee scoring rules} due to the design of our generic algorithm which expects a \textit{scoring function}. Chamberlin-Courant rule \cite{chamberlin1983representative} and Random Ballot \cite{amar1984choosing} are two such rules that have been shown to promote PR. We also consider SNTV (although it doesn't promote PR) because it has been shown to favour minorities \cite{cox1994strategic}, which increases diversity. Our experiments suggest that rules that demonstrate PR and diversity manifest as excellent exploratory RL algorithms.

\section{Experiments}
\label{sec:experiments}
\subsection{Environments}

\begin{figure}[!htbp]
    \includegraphics[width=0.45\textwidth]{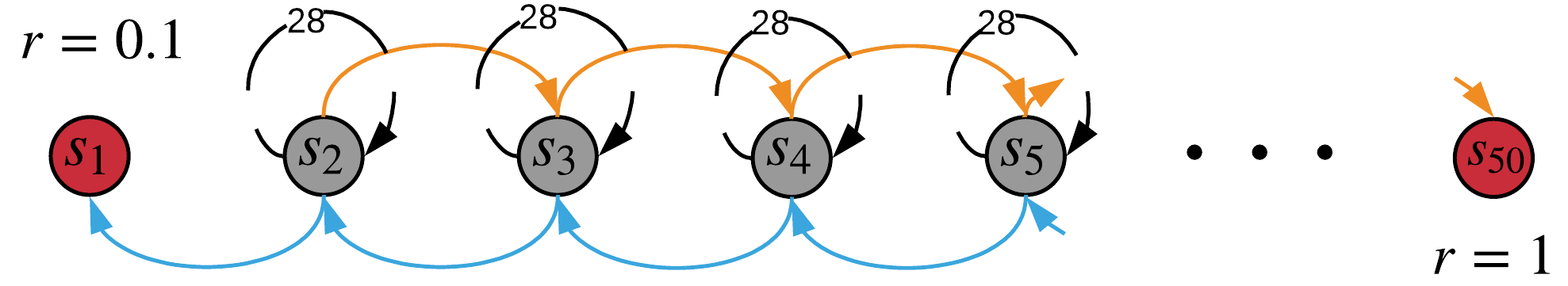}
    \centering
    \caption{\label{fig:corridor-mdp} Corridor MDP with 30 actions and 50 states.}
\end{figure}

The first evaluation was done on a combination of corridor MDP from Dueling Architectures \cite{2015arXiv151106581W} and a  MDP from Bootstrapped DQN \cite{NIPS2016_6501}. As shown in Figure \ref{fig:corridor-mdp}, this combined corridor MDP consists of a linear chain of $50$ states, out of which only two states, $s_1, s_{50}$, confer rewards. The start state is randomly selected based on a Binomial distribution biased towards the low rewarding state $s_0$, i.e. start state is $s_{i+1} | i \sim Binomial(n=49,p=0.2)$. Every non terminal state has $m$ actions, out of which randomly selected $m-2$ actions are non-operational while the other two leads to a transition to adjacent left and right states. Every episode has a length of $100$ steps after which it is terminated.

On five instances of this environment with increasing action-set sizes $m \in \{10,20,30,40,50\}$, we evaluated two sets of similar ensemble algorithms: (i) Bloc, Majority Voting, and SNTV Q-learning, and (ii) Borda, CCR, and Rank Voting Q-learning. The results are shown in Figure \ref{fig:comparision}. SNTV and Bloc Q-learning are multi-winner extensions to Majority Voting Q-learning using the dynamic resizing method discussed in Section \ref{sec:dyn-resizing}. Similarly, CCR and Borda Q-learning are multi-winner extensions to Rank Voting Q-learning. 

\begin{figure}[H]
    \begin{subfigure}[b]{0.2\textwidth}
        \centering
        \includegraphics[width=\textwidth]{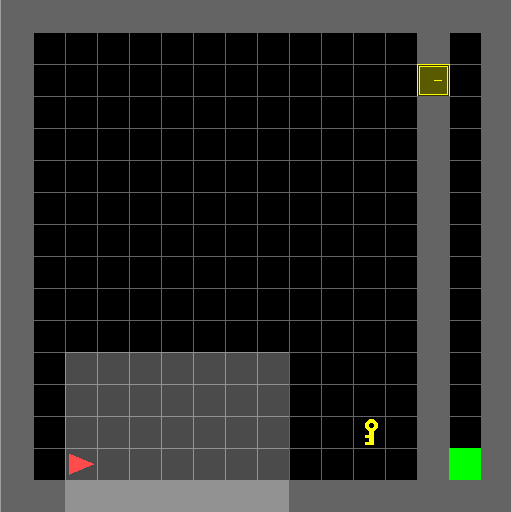}
        \subcaption{\label{fig:doorkey}DoorKey}
    \end{subfigure}
    \begin{subfigure}[b]{0.2\textwidth}
        \centering
        \includegraphics[width=\textwidth]{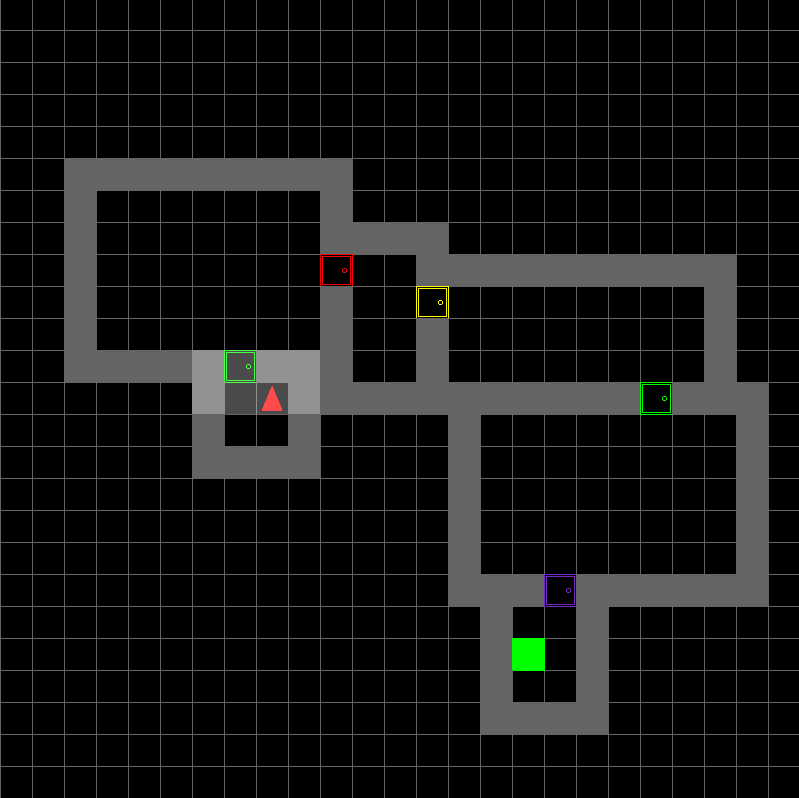}
        \subcaption{\label{fig:multiroom}MultiRoom}
    \end{subfigure}
    \begin{subfigure}[b]{0.2\textwidth}
        \centering
        \includegraphics[width=\textwidth]{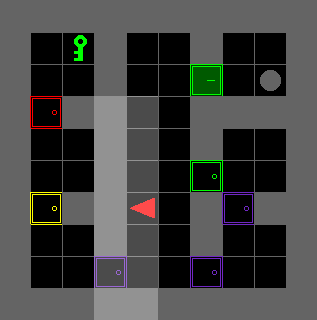}
        \subcaption{\label{fig:keycorridor}KeyCorridor}
    \end{subfigure}
    \begin{subfigure}[b]{0.2\textwidth}
        \centering
        \includegraphics[width=\textwidth]{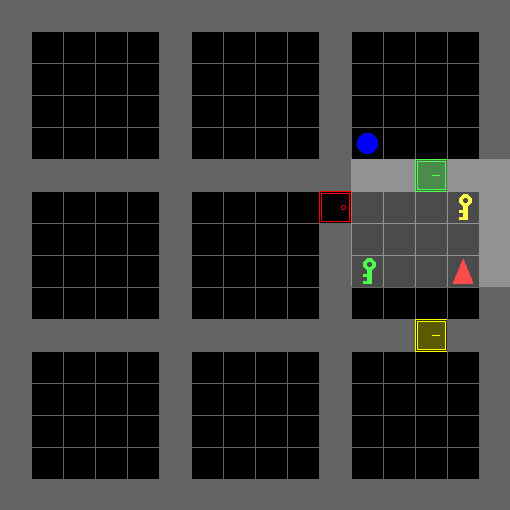}
        \subcaption{\label{fig:obstructedmaze}ObstructedMaze}
    \end{subfigure}
    \centering
    \caption{\label{fig:grid-world} Objective in Figure \ref{fig:doorkey} and \ref{fig:multiroom} is to reach the checkpoint (green area). In Figure \ref{fig:keycorridor} and \ref{fig:obstructedmaze}, the goal is to find the key (correct one) to the locked room and interact with the object (circle).}
\end{figure}

Next, we evaluated the ensemble algorithms on a test bed of grid-world puzzles \cite{gym_minigrid}, shown in Figure \ref{fig:grid-world}. The environments are partially observable, with varying objectives such as navigating to the goal, finding correct key, and unlocking the correct door. The action-set has $7$ discrete actions and the rewards in all of these environments are extremely sparse: the agent receives $1 - 0.9 \times (step\_count / max\_steps)$ on successful completion of task and zero otherwise. Several ensemble RL algorithms were trained on these environments for 2 million steps each across 48 different seeds (i.e. 48 random grid-world maps). For all the runs, the ensemble size was $k=10$, discount factor was 0.9, learning rate was 0.2, and the exploration schedule was linear (annealed from 1 to 0.001 in 1 million steps). The scoring function thresholds for multi-winner algorithms were manually tuned and set to $S_{thresh} = 10$ for SNTV and Bloc and to $S_{thresh} = 68$ for CCR and Borda.

\subsection{Results}
\label{sec:results}

\begin{figure*}[hbt!]
    \centering
    \begin{subfigure}[t]{\textwidth}
        \centering
        \includegraphics[width=1\textwidth]{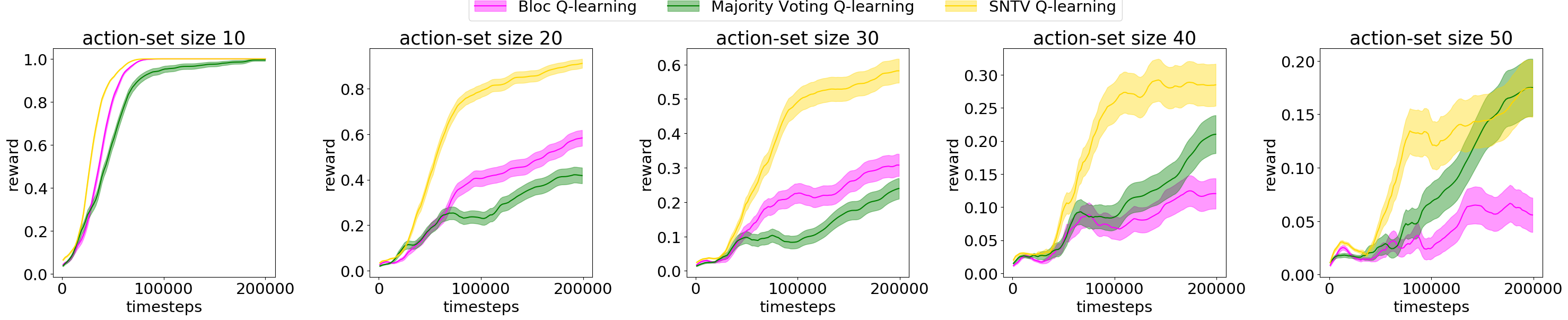}
        \subcaption{\label{fig:comparision-Bloc-SNTV} Rewards vs time-steps for Bloc, Majority Voting, and SNTV Q-learning.}
    \end{subfigure}
    \par\bigskip 
    \begin{subfigure}[t]{\textwidth}
        \centering
        \includegraphics[width=1\textwidth]{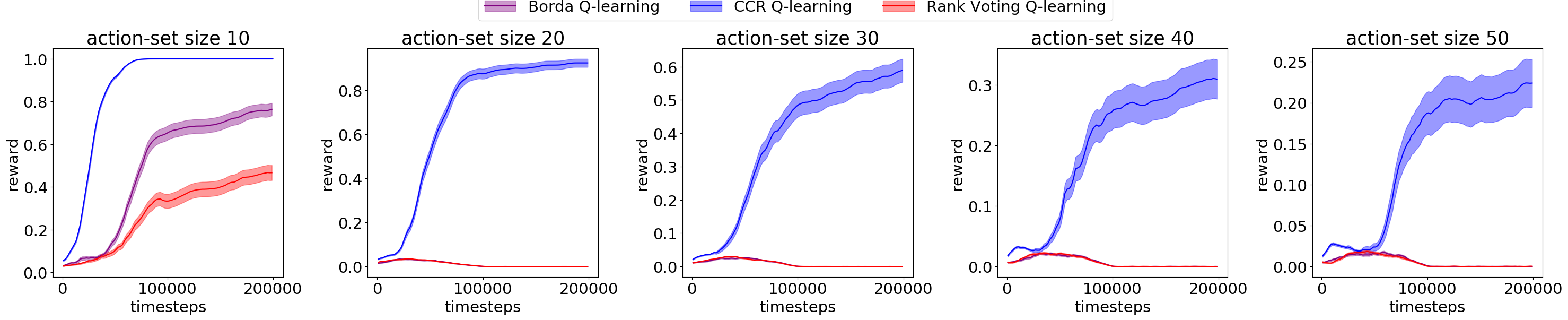}
        \subcaption{\label{fig:comparision-Broda-CCR} Rewards vs time-steps for Borda, CCR, and Rank Voting Q-learning}
    \end{subfigure}
    \caption{\label{fig:comparision} Comparison of ensemble Q-learning algorithms on corridor MDP for increasing action-set sizes.}
\end{figure*}

\begin{table*}[hbt!]
    \centering
    \resizebox{\textwidth}{!}{\begin{tabular}{|l|cccc|cccc|}
        \hline 
                                      & \multicolumn{4}{c|}{single-winner voting}  & \multicolumn{4}{c|}{multi-winner voting}      \\
                                      & Average & Rank & Majority & Lottery        & Bloc & Borda  & SNTV          & CCR           \\ 
        \hline
        MiniGrid-DoorKey-16x16-v0      & 0.08    & 0.06 & 0.37     & 0.38           & 0.41 & 0.11  & 0.49          & \textbf{0.58} \\
        MiniGrid-MultiRoom-N6-v0       & 0.00    & 0.01 & 0.13     & \textbf{0.47}  & 0.40 & 0.00  & \textbf{0.47} & \textbf{0.47} \\
        MiniGrid-KeyCorridorS4R3-v0    & 0.10    & 0.09 & 0.53     & 0.90           & 0.22 & 0.13  & \textbf{0.92} & \textbf{0.92} \\
        MiniGrid-ObstructedMaze-2Dl-v0 & 0.07    & 0.08 & 0.20     & 0.52           & 0.16 & 0.06  & 0.63          & \textbf{0.64} \\
        \hline
    \end{tabular}}
    \caption{\label{tab:result-comparision}Performance comparison (metric described in Section \ref{sec:results}) for different voting rules based RL algorithms on grid-world environments.}
\end{table*}

Figure \ref{fig:comparision} shows the results of the evaluation on corridor MDP. Table \ref{tab:result-comparision} lists the performance comparison of the evaluations on the grid-world environments. The metric used is the maximum value of exponential moving averages, meaned across 48 runs. In order to do this, we form exponential moving average estimates for each run and use it to sample 100 re-estimates at equidistant points (step difference of 20000). The samples are then meaned across runs and the maximum value is selected.



The results suggest that in general the multi-winner voting RL algorithms fare much better compared to single-winner variants, highlighting the efficacy of threshold based dynamic resizing method. Moreover, we see a consistent pattern of PR and diversity based algorithms---CCR, SNTV and Lottery Q-learning---beating other ensemble Q-learning techniques and being more resilient to increasing exploration difficulty. We believe this provides an alternate reasoning as to why Bootstrapped DQN \cite{NIPS2016_6501} does deep exploration: the random sampling of heads is akin to Lottery voting rule which exhibits PR.

\section{Concluding Discussion}

In this paper, we presented a committee voting based unified aggregation mechanism that generalizes several ensemble based Q-learning algorithms. By proposing a dynamic resizing committee election mechanism, we extended classical Q-learning to a generic ensemble RL algorithm. On plugging a multi-winner committee voting rule in this generic procedure, we see that the resulting RL algorithm manifests the underlying voting rule's properties. For instance, \textit{proportional representation} centric voting rules such as Chamberlin-Courant and Random Ballot exhibit an improvement in exploration on use with the generic algorithm, as seen in our experiments on fabricated MDPs as well as complex grid world environments.  

While our analysis focused only on exploratory behaviour of ensemble RL algorithms, one may investigate other properties, such as stability to environmental perturbation via application of \textit{Gehrlein Stable} voting rules \cite{gehrlein1985condorcet,aziz2017condorcet} such as Minimal Size of External Opposition rule (SEO) and Minimal Number of External Defeats (NED) \cite{coelho2005understanding}. Several other multi-winner voting rules could potentially be of interest in modifying RL algorithm's traits and our work provides a method to study them.

\label{sec:conclusion}



{
\bibliographystyle{named}
\bibliography{osrl}
}

\clearpage
\appendix
\section{Appendix}

\subsection{Proof of Proposition \ref{prop:equivalence}}


From the definitions of voting rules, we can see that in case when committee size $n = 1$, $f_{plurality}$ \eqref{eq:plurality} is identical to $f_{bloc}$ \eqref{eq:bloc} and $f_{ccr}$ \eqref{eq:chamberlin-courant} is identical to $f_{borda}$ \eqref{eq:broda}. Therefore showing equivalence for either of the rule in a pair is sufficient.

For plurality, this equivalence is established as follows.
	\begin{equation}
		\label{eq:mv-election}
		\begin{aligned}
			\pi^{MV}_t(s_t) 	
			                &= \argmax_{a \in A} \sum_{i \in K} \mathbbm{1} \{ a = \argmax_{a' \in A} Q^i_t(s_t, a') \} \\
							&= \argmax_{a \in A} \sum_{i \in K} \mathbbm{1} \{ a = \argmax_{a' \in A} \mu^i(a') \} \\
							&= \argmax_{a \in A} \sum_{i \in K} \mathbbm{1} \{ pos^i(a) = 1 \} \\
							&= \argmax_{a \in A} \sum_{i \in K} \alpha^i_1(a) \\
							&= \argmax_{a \in A} \sum_{i \in K} f_{plurality}(\mu^i, \{a\}) \\
							&= \mathcal{M}_{f_{plurality}}(E, 1)
		\end{aligned}
	\end{equation}
The proof for remaining rules follows a similar flow.

\subsection{Proof of Proposition \ref{prop:equivalence-boltzmann}}



Let $g(a) = \sum_{i \in K} \mu^i(a)$. Let the $l$-sized winning committee $\mathcal{M}_{f} (E, l)$ be denoted as $\mathcal{M}_l$ and the winning score $\mathcal{G}_{f} (E, l)$  as $\mathcal{G}_l$. We can express $\mathcal{G}_l$ as

\begin{equation}
    \label{eqn:judge-score-additive}
    \mathcal{G}_l = \sum_{a \in \mathcal{M}_{f} (E, l)} g(a).
\end{equation}



A voting rule is a \textbf{best-k rule} if there exists a preference function $h$ such that for each election $E$, the result $\mathcal{M}_{f} (E, l)$ for any committee-size $l \in [n]$ is same as top $l$ ranking candidates in $h$. One may easily verify using \eqref{eqn:judge-score-additive} that for majority judge voting rule, the preference function is $h = g$, and therefore is a \textit{best-k} rule. \citet{DBLP:journals/corr/ElkindFSS15} showed that all \textit{best-k} voting rules follow \textit{committee monotonousity} and vice versa. Coupled with tie-breaking constraint, we therefore ensure that $\mathcal{M}_{f} (E, l-1) \subset \mathcal{M}_{f} (E, l)$ is true.

Let the extra element be $a_l$. The difference in scores is 
\begin{equation}
    \begin{aligned}
    \mathcal{G}_l - \mathcal{G}_{l-1} &= \sum_{a \in \mathcal{M}_{f} (E, l)} g(a) - \sum_{a \in \mathcal{M}_{f} (E, l-1)} g(a) \\
                                      &= g(a_l) = \sum_{i \in K} \mu^i(a) \\
                                      &= \sum_{i \in K} \frac{e^{Q^i_t(s,a_l)}}{\sum_{a \in A} e^{Q^i_t(s,a)}} \\
                                      &= \abs{K} \times \pi^{BA}_t(a_l|s)
    \end{aligned}
\end{equation}

\subsection{Proof of Proposition \ref{prop:bootstrapped}}

Except the inclusion of bootstrapping mask, the proof follows along the lines of proposition \ref{prop:equivalence}.




\end{document}